\documentclass[11pt,a4paper]{article}
\usepackage[hyperref]{acl2020}
\usepackage{times}
\usepackage{latexsym}
\usepackage{graphicx}

\usepackage{amsfonts}

\usepackage{microtype}

\aclfinalcopy

\title{Do Transformers Need Deep Long-Range Memory?}

\author{Jack W. Rae \\
  DeepMind \& UCL \\
  London, UK \\
  \texttt{jwrae@google.com} \\
  \And 
  Ali Razavi \\
  DeepMind \\
  London, UK \\
  \texttt{alirazavi@google.com}
}

\date{}

\begin{document}
\maketitle
\begin{abstract}
Deep attention models have advanced the modelling of sequential data across many domains. For language modelling in particular, the Transformer-XL --- a Transformer augmented with a long-range memory of past activations --- has been shown to be state-of-the-art across a variety of well-studied benchmarks. The Transformer-XL incorporates a long-range memory at every layer of the network, which renders its state to be thousands of times larger than RNN predecessors. However it is unclear whether this is necessary. We perform a set of interventions to show that comparable performance can be obtained with 6X fewer long range memories and better performance can be obtained by limiting the range of attention in lower layers of the network.
\end{abstract}
\section{Introduction}
When we read a book, we maintain representations of the characters and events in the text that help us understand the story. We do this with a selective memorisation process; most of the finer details of the text are quickly forgotten and we retain a relatively compact representation of the book's details. 

Early models of natural language used recurrent neural networks (RNNs) such as the Long Short-Term Memory \citep{hochreiter1997long} which emulated this selective memory approach by modelling the past in a compact state vector. The model learns to store relevant information within its state implicitly in order to optimise the task loss. 

The LSTM has reigned as a state-of-the-art language model for over two decades since its inception in the '90s \citep{melis2017state} and is arguably the most ubiquitous neural sequence model. Unlike human memory systems, however, the LSTM struggles to reason over long-range contexts when reading text. This has been observed in multiple contexts. In the carefully curated LAMBADA benchmark \citep{paperno-etal-2016-lambada} which tests language model predictions on sections of book text that have long term structure as decided by human raters, LSTMs completely fail. Namely LSTMs guess the correct word $0\%$ of the time, where humans are considered to be above $70\%$ accuracy. For regular language modelling, \citet{daniluk2017frustratingly} observed that an LSTM augmented with attention would rarely attend beyond seven preceding words of context. Samples from LSTMs language models quickly devolve into generic text devoid of an overall theme. This has lead many to wonder whether there is any non-negligible long-range signal in the task of language modelling.

Recently we have seen that deep attention models can draw long-range signal from text, even when the objective is as simple as next-word prediction. With the advent of the Transformer \citep{vaswani2017attention}, significant gains in language modelling performance can be obtained by extending the models' attention to thousands of words. The Transformer-XL \citep{dai2019transformer}, a Transformer variant specialised for long-range sequence modelling via the introduction of a cache of past activations, obtained state-of-the-art results in the four major LM benchmarks --- PTB \citep{mikolov2010recurrent}, LM1B \citep{chelba2013one}, Enwik8 \citep{hutter2012human}, and WikiText \citep{merity2016pointer}. In the case of the latter two, \citet{dai2019transformer} showed the model effectively used over one thousand words of context, and the resulting samples reflect a thematic consistency spanning paragraphs. When Transformers are paired with long contexts and a large amount of data, e.g. GPT-2 \citep{radford2019language} and Megatron \citep{shoeybi2019megatronlm}, the resulting samples are remarkable in their long-range consistency and stylistic realism.  

However Transformers abandon the compact and selective representation of the past. They store a hidden activation at every time-step (up to a given attention range) and every layer within the network. This can consume orders of magnitude more space than prior RNN hidden states, or the original text. E.g. a typical state-of-the-art LSTM language model state size may range from 4KB \cite{rae2018fast} to model Wikipedia articles to 64KB \citep{jozefowicz2016exploring} to model news --- and is never greater than 1MB. Whereas a current state-of-the-art 18-layer Transformer-XL state size for Wikipedia articles is 112MB. The state is so large because a separate memory (e.g. 1600 vectors of size d=1024) is maintained per layer. If this were found to be unnecessary then we can reduce the state's memory considerably.

In this paper we investigate a simple question: can we use short-range attention for the majority of layers in the Transformer and recover the same performance? The hypothesis is that this should be possible, because many steps of reasoning will only involve short-range correlations, i.e. to piece characters together to form words or phrases.
We find indeed it is possible. We  recover comparable performance for long-range language modelling by using a small fraction (1/6th) of long-range memories to the baseline TransformerXL. Crucially, we find it matters \textit{where} long-range memories are placed in the network. Placing them in the lower layers of the network is ineffective; placing them in the latter layers or interleaved across the network works much better. We show that such a model trains with $2X$ less time and memory, due to the reduction in expensive attention operations.

\section{Background}
The \textit{Transformer} is a deep neural network for processing sequences \citep{vaswani2017attention}, it processes a window of $n$ consecutive inputs $x_{t-n}, \ldots, x_t$ in parallel. At each layer it reasons over time using \textit{multi-head attention} which we will briefly describe. For a given layer $l$, let $h_t \in \mathbb{R}^{1 \times d}$ be the hidden activation at time $t$, and $h_{\le t} \in \mathbb{R}^{t \times d}$ be the preceding activations in the same window. Let $k$ be the number of attention heads, then $Q_i, K_i, V_i \in \mathbb{R}^{d \times \frac{d}{k}}$ are a set of learnable weight matrices which generate \textit{queries}, \textit{keys}, and \textit{values} per attention head. These are defined to be $q_i = h_t Q_i$ as the query, $k_i = h_{\le t} K_i$ to be the keys, and $v_i = h_{\le t} V_i$ to be the values for attention head $i$. The attention head output is defined to be,
\[ attn_i(h_t, h_{\le t}) = \sigma(q_i k_i^T)v_i\]
where $\sigma(\cdot)$ is defined to be the softmax operator. Attention is the linear combination of each attention head, $attn = \sum_{i = 1}^k W_i \, attn_i$ with a learnable weight.

The attention operation consumes $\mathcal{O}(n)$ compute per step and thus $\mathcal{O}(n^2)$ for the window of inputs at each layer. The \textit{Transformer-XL} (TXL) proposes concatenating the past activations from the same window $h_{\le t}$ with a memory of size $m \ge n$ of past activations from the preceding windows of inputs \citep{dai2019transformer}. This results in an attention cost of $\mathcal{O}(n(n+m))$ which can be significantly cheaper than processing all $n + m$ inputs in parallel, which would require $\mathcal{O}((n+m)^2)$. The TXL's memory can be considered to be a state, alike to an RNN. However it requires a considerable space: $l \times m \times d$. For character-level language modelling \citet{dai2019transformer} use a 24-layer model on Enwik8, with memory size $m = 3800$, and hidden size $d = 1024$; this consumes 356MB at single precision. In contrast, the average article size is 8KB.
\begin{figure}
    \centering
    \includegraphics[width=0.98\linewidth]{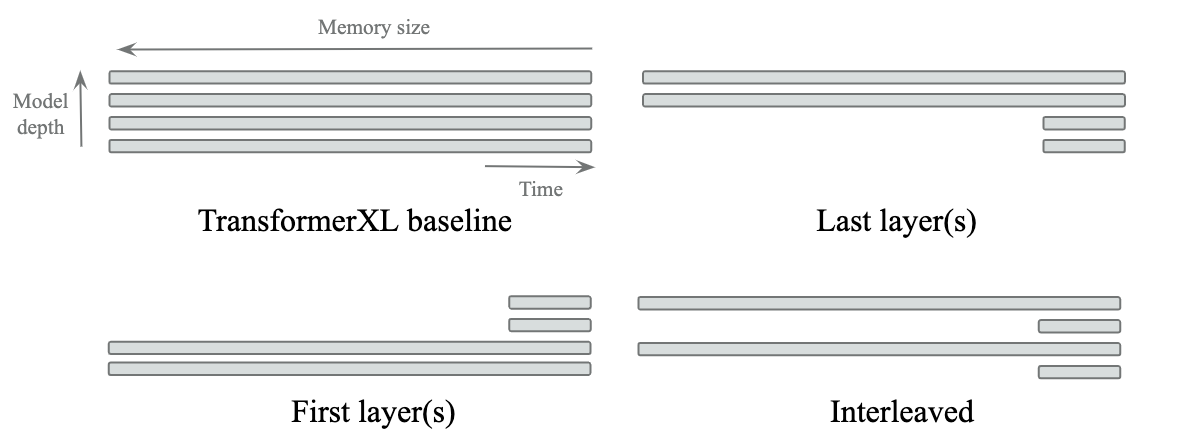}
    \caption{Comparison of arrangement patterns for long-range and short-range memories across the layers of a Transformer. Baseline contains equally long-range memories at every layer. }
    
    \label{fig:model}
\end{figure}
\section{Experiments}
We investigate whether the Transformer-XL can perform comparably with fewer long-range memory (LRM) layers on the two prominent long-range language modelling benchmarks, Enwik8 and WikiText-103.
\subsection{Interventions}
We perform intervention experiments where we replace the long-range memory, for a given layer, with a short-range memory (SRM) of size $m_{s} = 128$ for a subset of layers. We choose $m_s = 128$ because the TPUv3 contains a 128x128 matrix multiply unit, and any smaller size (other than zero) is padded up to 128. Thus it is a reasonable small size. We chose $m_s > 0$ such that the oldest activations have some context. Because we only modify the memory sizes of the model, which are independent of parameter count, \textbf{the number of model parameters is always held constant} (277M for Enwik8 and 257M for WikiText-103).

We consider a model with a varying number of LRMs from $l$ (the number of layers in the network, i.e. the usual case) to a range of fewer values, $\frac{l}{2}$, $\frac{l}{6}$, $1$, and $0$. We also consider where the LRMs should be arranged within the network; considering (i) interleaved with equal spacing, (ii) the first layer(s) of the network, and (iii) the latter layer(s) of the network. This is displayed visually in Figure \ref{fig:model}. 

\subsection{Model Setup}
Aside from memory configurations, we use an identical model setup to \citet{dai2019transformer}. During training we periodically evaluate on the validation set to choose an early stopping criterion. In the case of Enwik8 we periodically evaluate on the first 500K characters of the validation set to speed up model evaluation.
We train all models with an overall batch size of $32$, using 16 TPUv3 chips running synchronously. We use a window size of $n = 384$, a long-range memory (LRM) size of $m = 2304$. At test-time we extend the LRM size to $m=6000$, chosen from a sweep over the validation set. 

\section{Results}
\label{sec:results}
\begin{figure}[]
    \centering
    \includegraphics[width=0.8\linewidth]{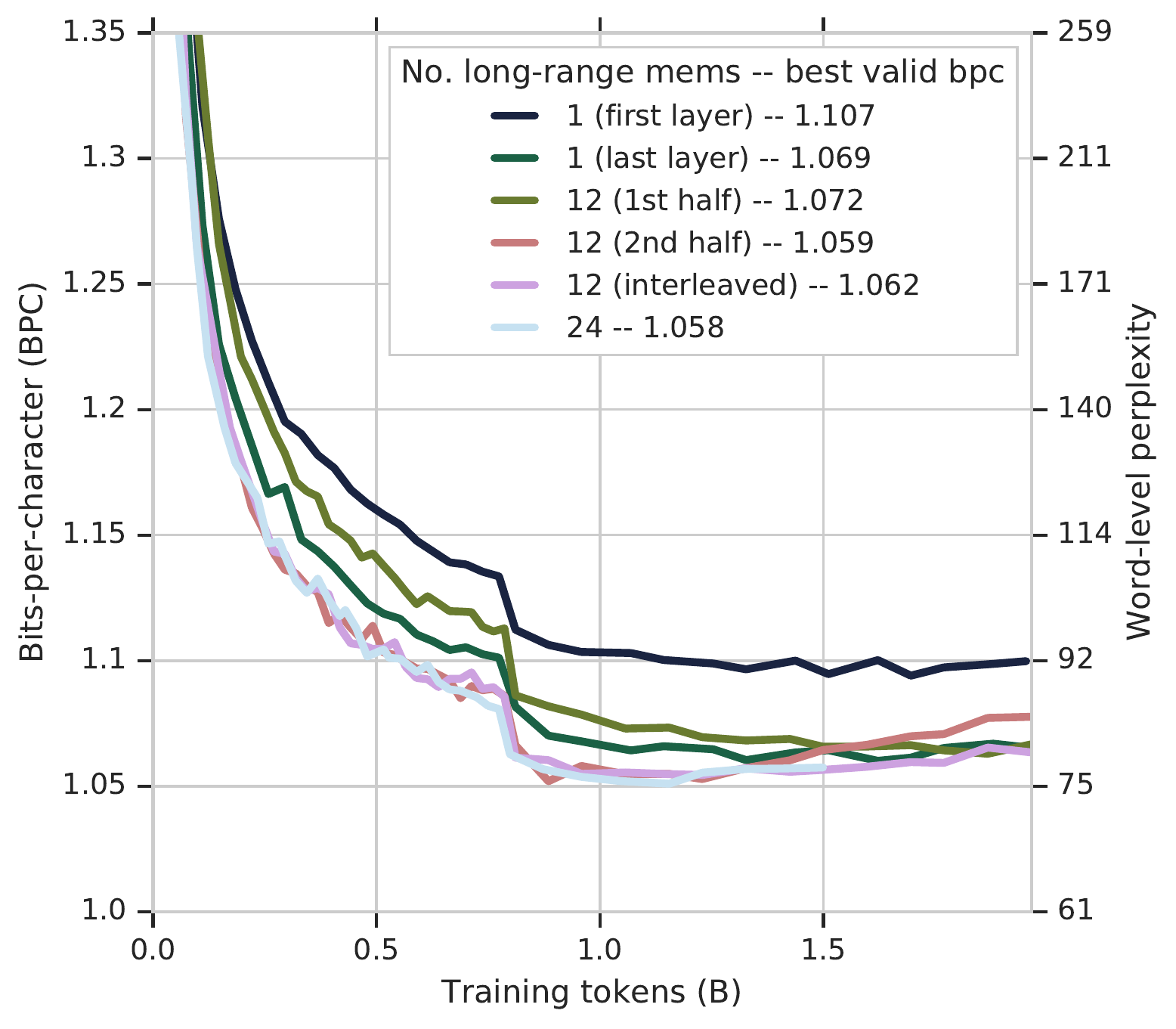}
    \caption{Enwik8 learning curves for varying long-range memory arrangements and no. layers. BPC over the first 500K characters from validation.}
    \label{fig:valid}
\end{figure}
We plot the \textbf{Enwik8} learning curves for a subset of layer variants in Figure \ref{fig:valid}. The worst-performing, is the variant with a single long-term memory at the lowest layer (black curve). However perhaps more surprisingly, we see a model with 12 LRMs at the lower layers of the network is actually \textit{worse} than a model with a single LRM on the final layer (dark green). We then see that the full TXL with 24 LRMs is seemingly identical to the 12 LRM models, with either LRMs interleaved across the whole model or LRMs placed in the final 12 layers. Note, we were not able to run these models with multiple seeds per hyper-parameter configuration - but we do generally find language models optimise consistently (e.g. unlike deep reinforcement learning models).
\begin{figure}
    \centering
    \includegraphics[width=0.94\linewidth]{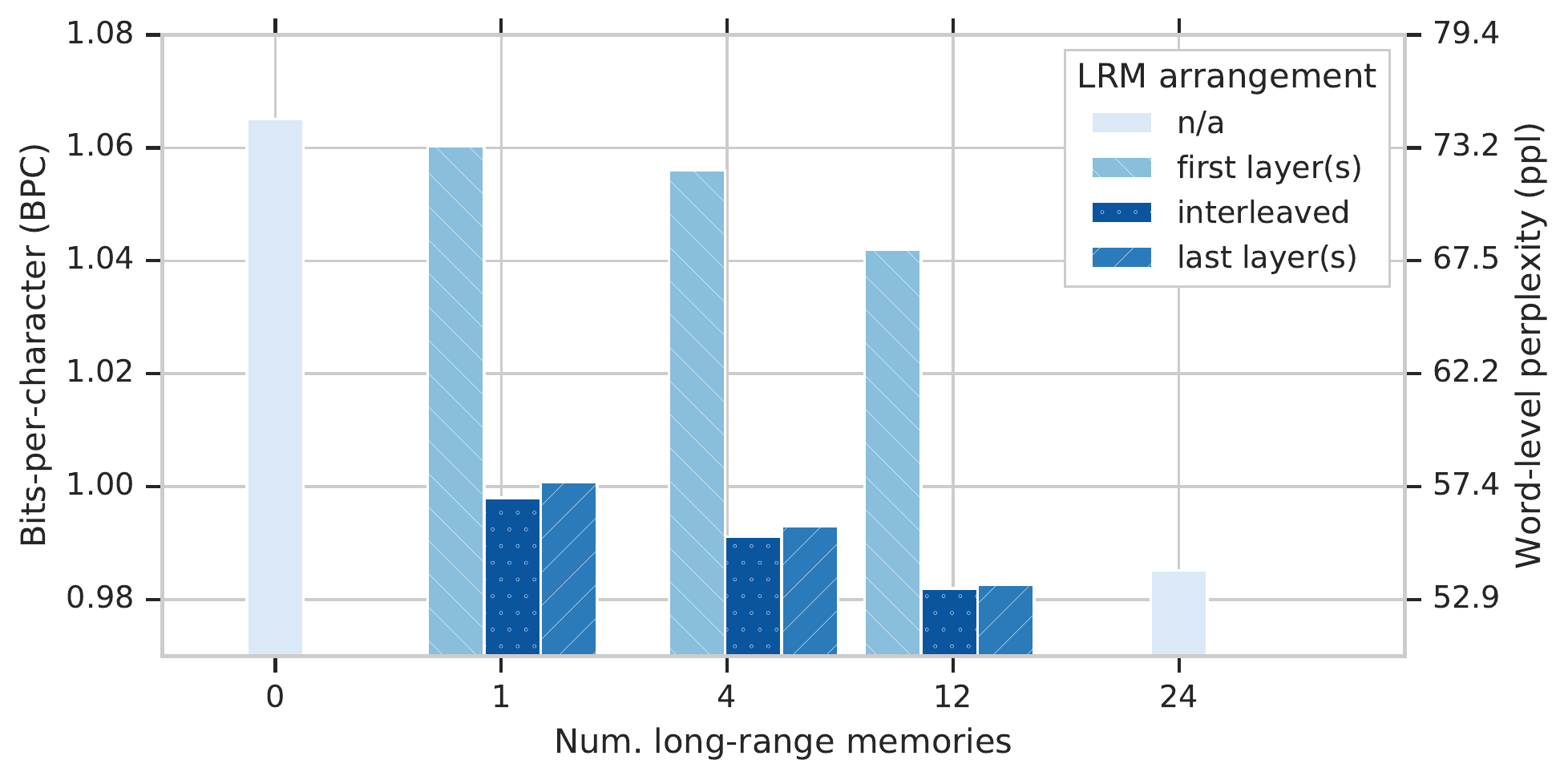}
    \caption{Enwik8 test performance over a varying number of long-range memories and arrangement patterns. Lower is better. Model: 24-layer Transformer-XL, evaluation long-range memory size: $6000$ (trained with $2304$) and short-range memories size: $128$.}
    \label{fig:test_bar}
\end{figure}

We show the final test performance in bits-per-character (BPC) alongside the corresponding word-level perplexity for models with a varying number of LRMs and LRM arrangements in Figure \ref{fig:test_bar}. Position clearly matters, if we place long-range memories in the first layers then performance is significantly worse. We hypothesise that this is because it is better to build up representations with local context before exploiting long-range correlations. For example, we need to piece together characters into an identified named entity (say) before we should query thousands of time-steps back for its prior occurrence.

We followed-up by running an additional arrangement of only placing LRMs in the \textit{middle} layers and found this to be worse than interleaved or final ($1.01$bpc for 4 long-range memories) which shows there is significant benefit to having some long-range memories in the higher layers.

Crucially, we are able to match (and slightly exceed) the full model's test performance with 12 LRMs, and even a model with 4 LRMs is very close ($\mathbf{0.9846}$ w/ 24 vs $\mathbf{0.9916}$ w/ 4 interleaved). It is worth noting that our TXL baseline actually outperforms the published version on Enwik8: $0.985$ BPC (ours) vs $0.993$ \citep{dai2019transformer}, which provides credence to the quality of the experimental setup.

We also inspect word-level language modelling on \textbf{WikiText-103}, using the same $18$-layer TransformerXL parameters \citep{dai2019transformer}. We obtain a baseline test perplexity of $18.3$ (matching the published value), and obtain \textbf{18.4} and \textbf{18.6} for interleaved and last-layer spacing respectively when using $l/6$ (i.e. 3) LRMs. We also try placing 3 LRMs on the first three layers and obtain 20.1 perplexity. We remark that (i) long-range memory is important for a significant improvement in performance, (ii) it is better to not place LRMs in the shallow layers, and (iii) it is not necessary to have as many long-range memories as model-layers for comparable modelling performance.

\subsection{Performance}
We show the performance of training the Transformer-XL with a varying number of LRMs for the Enwik8 architecture in Table \ref{tab:runtime}.
This shows the latency (per input token) and peak activation memory consumption during a training iteration on Enwik8 for a range of long-range memory layers.
We see the reduction of long-range memories from $24$ layers to $4$ layers cuts the activation peak memory by 3X. Thus it can be a worthwhile and simple performance improvement.
\begin{table}[]
    \centering
    \begin{tabular}{cc c}
    Num. LRMs   & Memory (GB)    & Time / token (us)  \\
    \hline
    24              &     3.4       & 405 \\
    12              &     2.8       & 273  \\
    4               &     1.1       & 191 \\
    1               &     0.50      & 155 \\
    0               &     0.20      & 143 \\
    \end{tabular}
    \caption{Profiling a 24-layer TXL training on Enwik8.}
    \label{tab:runtime}
\end{table}
\subsection{Varying Short-Range Memory}
In the preceding experiments we fix the short-range memory (SRM) length to $128$ and vary the frequency and arrangement of long-range memory layers. We now consider varying the length of SRM for an architecture with $\frac{l}{6}$ long-range memories to determine whether this impacts modelling performance.

\begin{figure}
    \centering
    \includegraphics[width=0.94\linewidth]{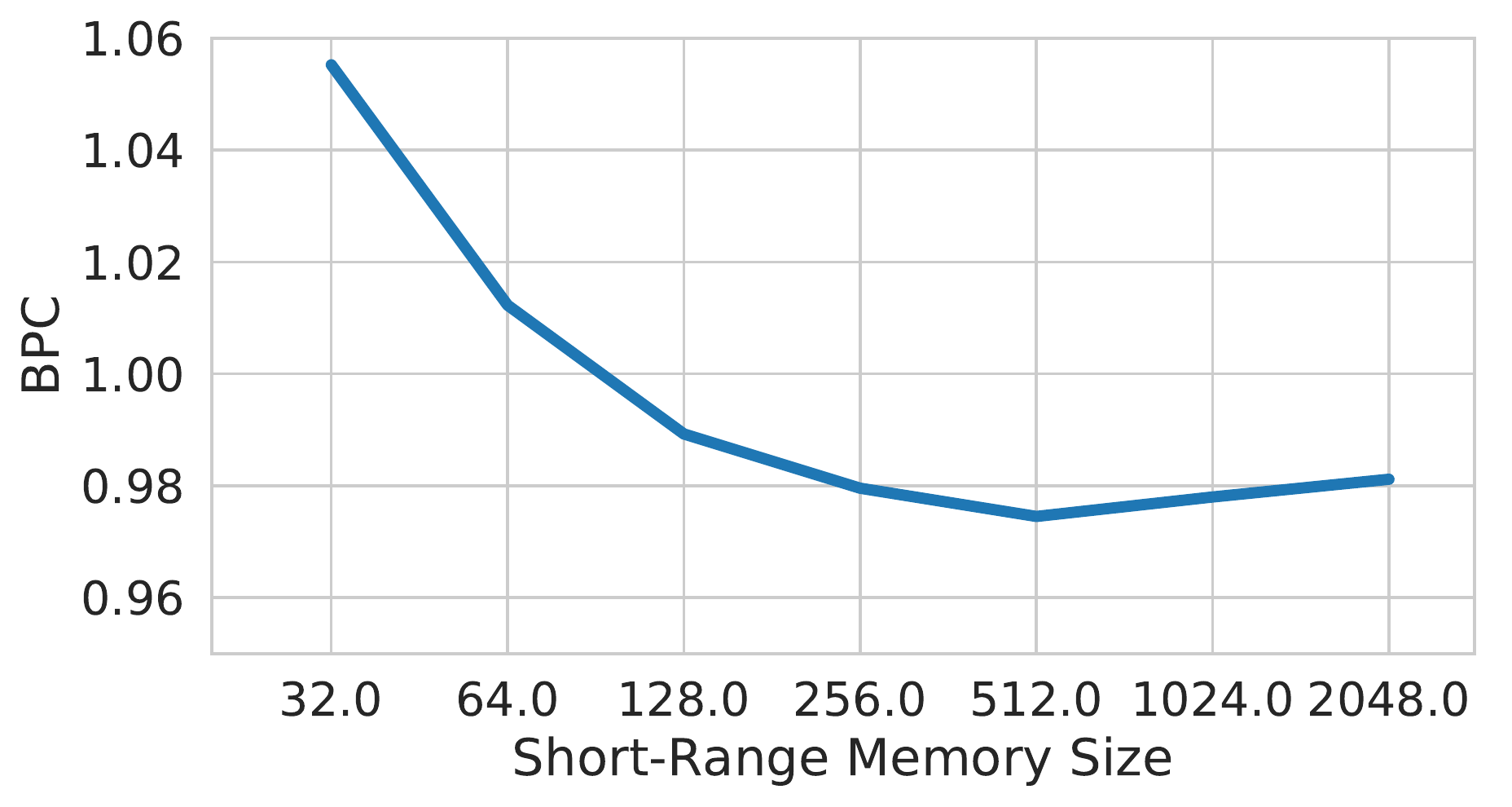}
    \caption{Enwik8 test performance for varying short-range memory length (at both train and test). TransformerXL model uses 4 interleaved long-range memories (trained 2304, tested 6000) and 20 short-range memory layers.}
    \label{fig:srm_sweep}
\end{figure}

We train (and evaluate) the model with twenty SRM lengths from 32-2048, and incorporate four interleaved LRM layers (trained at 2304, evaluated at 6000). The results are plotted in Figure~\ref{fig:srm_sweep}. Shortening the memory size to less than 128 provides no speedup for our TPU training setup, as matrices are multiplied in 128x128 blocks, however it incurs a drop in modelling performance. Furthermore increasing the memory size beyond 512 further slows the model down and reduces modelling performance. We see an optimal SRM length is around 512 steps which obtains \textbf{0.974}BPC on Enwik8 --- a non-trivial performance boost over the 0.99BPC TransformerXL baseline. Thus we conclude that limiting the range of attention can not only speed up the model but improve performance. 

\section{Related Work}
There have been several recent works exploring deep sequence models with a small attention window per layer. \citet{wu2019pay} proposed the \textit{dynamic convolution}, where the model directly produces a set of weights over a sequence in memory and then combines them with a convolution. The attention window is thus restricted to the convolution kernel size --- a couple of words. \citet{wu2019pay} show comparable performance to the Transformer at sentence-level machine translation. However they do not investigate longer-context applications.

\citet{rae2019compressive} propose shortening the range of attention for Transformers by compressing the distant past. They find the first layers of the model are the most compressible, and obtain state-of-the-art in several long-range language model benchmarks (WikiText-103 and Enwik8). However they do not consider restricting the range of attention for a subset of layers to save compute and space.
\citet{sukhbaatar2019adaptive} propose an \textit{adaptive attention} scheme for the TransformerXL where the model can learn to modulate the size of its attention window per attention head. They observe the neural network converges to using smaller attention spans for lower layers in the network, which adds additional evidence to the finding that long-range memories are not useful in these lower layers. Because \citet{sukhbaatar2019adaptive} place the range of attention in the optimisation problem it is very flexible. In this study we promote interpretability by making a set of direct interventions to the memory size across layers. This does result in less generality, as we explicitly create two types of attention ranges, where adaptive attention can select many. However ultimately the two approaches of generality and interpretability complement one another.

\cite{fan2020accessing} show that one can train a transformer by having all layers attend to a single memory that is the linear combination of all layers' memories. Thus at training all layers' memories are maintained, but at evaluation or generation time there can be a single memory. This gives evidence that we do not need to store many separate representations for long-range memory to perform well at test time, but the approach does require storing them during training --- and incurs significant slowdown to the model.

\section{Discussion}
We explore a set of interventions to the Transformer-XL's architecture that are very simple to implement, i.e. a few lines of code, but shed light on the fundamental workings of the model when modelling long sequences of text. In our set of interventions, we only modify the flow of information within the network, versus the number of trainable parameters. Thus we do not have confounding factors of varying network capacity.

Our finding is that we do not need long-range memories at every layer of the network. Comparable performance can be obtained with a fraction (1/6th) of long-range memories if they are spaced equally across the network, or in the latter layers. We hypothesise this is because modelling long-range correlations is best done when representations are first formed from short-range correlations. We also find a real performance drop using a single long-range memory, proving long-range dependency is not superfluous to the task.

This study has implications for practitioners interested in speeding up deep Transformer-XL models. There have been a number of long-range transformer variants published in the past year~\citep{lample2019large, rae2019compressive, roy2020efficient, kitaev2020reformer} which aim to extend the range of attention via sparsity or compression. However these models maintain the use of uniform memory capacity for each layer. Here we show that long-range attention does not need to be scaled for every layer, and thus these architectures can be further sped-up with this observation.

This study also has implications for researchers using a single long-range memory, which has typically been the approach in traditional RNN + attention systems. For example, the Differentiable Neural Computer \citep{graves2016hybrid} and recent memory-augmented agents for reinforcement learning, which utilise a distinct working memory with a single long-range episodic memory \citep{fortunato2019generalization}. Perhaps performance could be improved by adding additional layers of \textit{episodic} memories. 

The practice of storing deep long-range memories is not scalable if we wish for neural networks to have the kinds of large-horizon reasoning that humans possess. We believe the solution of maintaining a small number of long-range memories is a step towards tractable lifelong memory.

\bibliography{acl2020}
\bibliographystyle{acl_natbib}

\end{document}